\pdfoutput=1
\documentclass{article}

\usepackage{microtype}
\usepackage{graphicx}
\usepackage{subfigure}
\usepackage{booktabs} % for professional tables

\usepackage{hyperref}
\usepackage{url}
\usepackage{graphicx}
\usepackage{makecell}
\usepackage{multirow}
\usepackage{enumitem}
\usepackage{bbm}
\usepackage{color,xcolor}
\usepackage{booktabs}
\usepackage{lineno}
\usepackage{array}
\usepackage{ulem}
\usepackage{makecell}
\usepackage{multicol}
\usepackage{amsmath,amsthm,amsfonts,amssymb,bm}
\usepackage{stfloats}
\usepackage[font=small,labelfont=bf]{caption}
\usepackage{tabulary}
\newcolumntype{C}[1]{>{\centering\arraybackslash}p{#1}}
\newcolumntype{L}[1]{>{\arraybackslash}p{#1}}
\DeclareMathOperator*{\argmax}{arg\,max}

\usepackage[accepted]{icml2021}

\begin{document}

\twocolumn[
\icmltitle{SemiRetro: Semi-template framework boosts deep retrosynthesis prediction}



\icmlsetsymbol{equal}{*}

\begin{icmlauthorlist}
    \icmlauthor{Zhangyang Gao}{equal,to,goo}
    \icmlauthor{Cheng Tan}{equal,to,goo}
    \icmlauthor{Lirong Wu}{to,goo}
    \icmlauthor{Stan Z. Li}{to}
\end{icmlauthorlist}

\icmlaffiliation{to}{AI Research and Innovation Lab, Westlake University}
\icmlaffiliation{goo}{Zhejiang University}

\icmlcorrespondingauthor{Stan Z. Li}{Stan.ZQ.Li@westlake.edu.cn}

\icmlkeywords{Deep Learning, Graph, Retrosynthesis}

\vskip 0.1in
]



\printAffiliationsAndNotice{\icmlEqualContribution} % otherwise use the standard text.

\begin{abstract}
    Recently, template-based (TB) and template-free (TF) molecule graph learning methods have shown promising results to retrosynthesis. TB methods are more accurate using pre-encoded reaction templates, and TF methods are more scalable by decomposing retrosynthesis into subproblems, i.e., center identification and synthon completion. To combine both advantages of TB and TF, we suggest breaking a full-template into several semi-templates and embedding them into the two-step TF framework. Since many semi-templates are reduplicative, the template redundancy can be reduced while the essential chemical knowledge is still preserved to facilitate synthon completion. We call our method SemiRetro, introduce a new GNN layer (DRGAT) to enhance center identification, and propose a novel self-correcting module to improve semi-template classification. Experimental results show that SemiRetro significantly outperforms both existing TB and TF methods. In scalability, SemiRetro covers 98.9\% data using 150 semi-templates, while previous template-based GLN requires 11,647 templates to cover 93.3\% data. In top-1 accuracy, SemiRetro exceeds template-free G2G 4.8\% (class known) and 6.0\% (class unknown). Besides, SemiRetro has better training efficiency than existing methods.
\end{abstract}

\section{Introduction}

Retrosynthesis prediction \citep{corey1969computer,corey1991logic} plays a crucial role in synthesis planning and drug discovery, which aims to infer possible reactants for synthesizing a target molecule. This problem is quite challenging due to the vast search space, multiple theoretically correct synthetic paths, and incomplete understanding of the reaction mechanism, thus requiring considerable expertise and experience. Fortunately, with the rapid accumulation of chemical data, machine learning is promising to solve this problem \citep{coley2018machine,segler2018planning}. In this paper, we focus on the single-step version: predicting the reactants of a chemical reaction from the given product.

Common deep-learning-based retrosynthesis works can be divided into template-based (TB) \citep{coley2017computer,segler2017neural,dai2019retrosynthesis,chen2021deep} and template-free (TF) \citep{liu2017retrosynthetic,karpov2019transformer, sacha2021molecule} methods. Generally, TB methods achieve high accuracy by leveraging reaction templates, which encode the molecular changes during the reaction. However, the usage of templates brings some shortcomings, such as high computation cost and incomplete rule coverage, limiting the scalability. To improve the scalability, a class of chemically inspired TF methods \citep{shi2020graph,NEURIPS2020_819f46e5} (see Fig.~\ref{fig:global_structure}) have achieved dramatical success, which decompose retrosynthesis into subproblems: i) \textit{center identification} and ii) \textit{synthon completion}. Center identification increases the model scalability by breaking down the target molecule into virtual synthons without utilizing templates. Synthon completion simplifies reactant generation by taking synthons as potential starting molecules, i.e., predicting residual molecules and attaching them to synthons to get reactants.  Although various TF methods have been proposed, the top-$k$ retrosynthesis accuracy remains poor. Can we find a more accurate way to predict potential reactants while keeping the scalability?

To address the aforementioned problem, we suggest combining the advantages of TB and TF approaches and propose a novel framework, namely SemiRetro. Specifically, we break a full-template into several simpler semi-templates and embed them into the two-step TF framework. As many semi-templates are reduplicative, the template redundancy can be reduced while the essential chemical knowledge is still preserved to facilitate synthon completion. And we propose a novel self-correcting module to improve the semi-template classification. Moreover, we introduce a directed relational graph attention (DRGAT) layer to extract more expressive molecular features to improve center identification accuracy. Finally, we combine the center identification and synthon completion modules in a unified framework to accomplish retrosynthesis predictions.

We evaluate the effectiveness of SemiRetro on the benchmark data set USPTO-50k, and compare it with recent state-of-the-art TB and TF methods. We show that SemiRetro significantly outperforms these methods. In scalability, SemiRetro covers 98.9\% of data using 150 semi-templates, while previous template-based GLN requires 11,647 templates to cover 93.3\% of data. In top-1 accuracy, SemiRetro exceeds template-free G2G 4.8\% (class known) and 6.0\% (class unknown). Owing to the semi-template, SemiRetro is more interpretable than template-free G2G and RetroXpert in synthon completion. Moreover, SemiRetro trains at least 6 times faster than G2G, RetroXpert, and GLN. All these results show that the proposed SemiRetro boosts the scalability and accuracy of deep retrosynthesis prediction.

\section{Related work}

\textbf{Template-based models} TB methods infer reactants from the product through shared chemical transformation patterns, namely reaction templates. These templates are either hand-crafted by human experts \citep{hartenfeller2011collection,szymkuc2016computer} or automatically extracted by algorithms \citep{coley2017prediction,law2009route}. For a product molecule, due to the vast search space, multiple qualified templates, and non-unique matching sites for each template, it is challenging to select and apply the proper template to generate chemically feasible reactants. To handle those challenges, \citep{coley2017computer} suggests sharing the same templates among similar products. \citep{segler2017neural,baylon2019enhancing} employ neural models for template selection with molecule fingerprint as input. GLN \citep{dai2019retrosynthesis} learns the joint distribution of templates and products by decomposing templates into pre-reaction and post-reaction parts and introducing logic variables to apply structure constraints. And LocalRetro \citep{chen2021deep} simplifies the template by removing its background, i.e., structures that do not change during the reaction. TB methods are interpretable and accurate because they embed rich chemical knowledge into the algorithm. However, these methods do not consider the partial template based on the synthons (introduced latter), and the vast space of templates and incomplete coverage severely limit their scalability.

\begin{figure*}[h]
    \centering
    \includegraphics[width=6in]{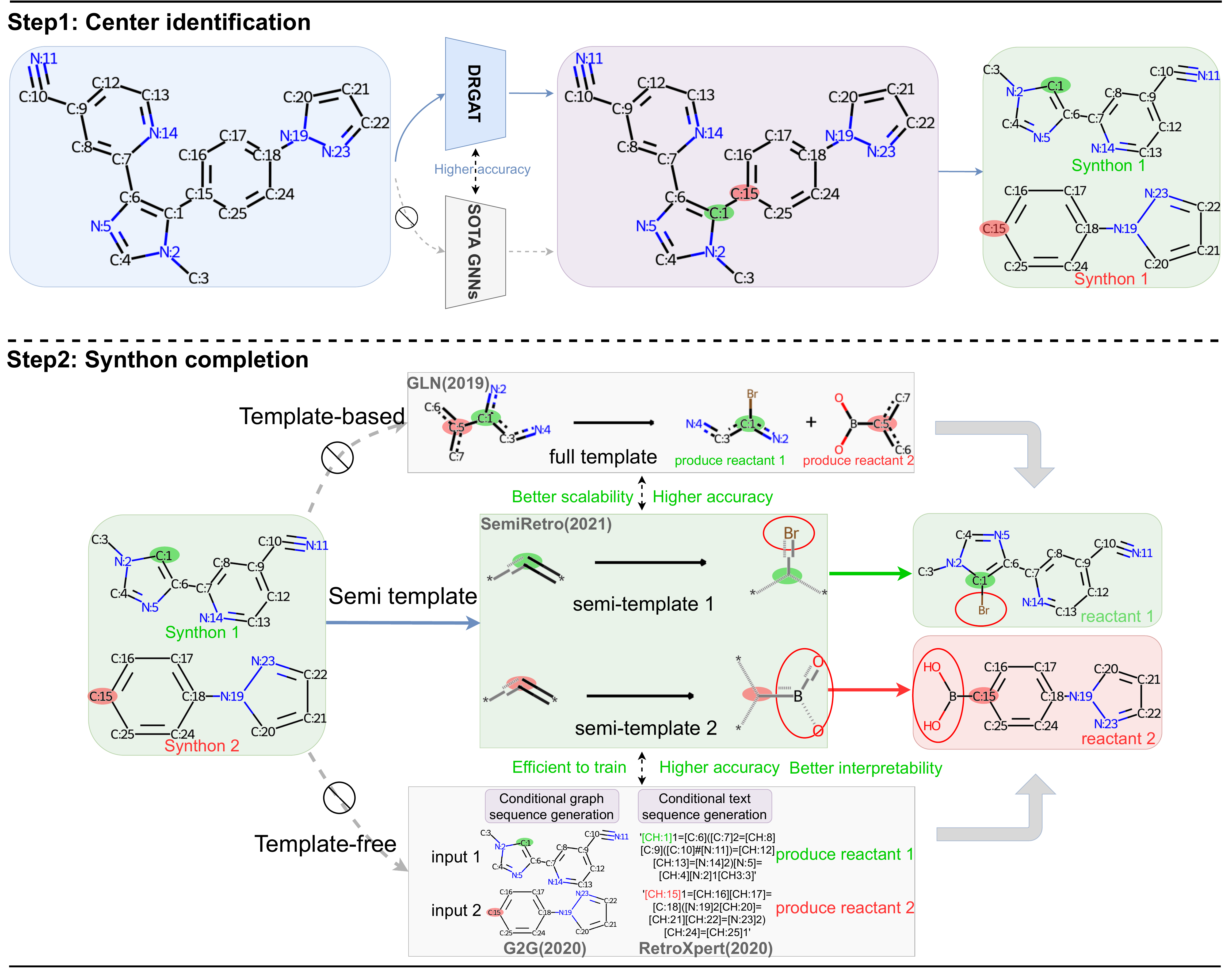}
    \vspace{-3mm}
    \caption{ Overview of SemiRetro. We decomposite retrosynthesis into two steps: \textit{center identification} and \textit{synthon completion}. In step 1, we use DRGAT to extract molecule features for predicting reaction centers. By breaking product bonds in these centers, synthons can be obtained.  In step 2, we use another DRGAT model to predict the semi-template for each synthon. The final reactants can be deduced from reaction centers, synthons, and semi-templates by using the residual attachment algorithm. }
    \label{fig:global_structure}
    \vspace{-3mm}
 \end{figure*}

\textbf{Template-free models} Instead of explicitly using templates, TF approaches learn chemical transformations by the model. \citep{liu2017retrosynthetic,karpov2019transformer} solve the retrosynthesis problem with seq2seq models, e.g. Transformer \citep{vaswani2017attention}, LSTM \citep{hochreiter1997long}, based on the SMILES representation of molecules. Despite the convenience of modeling, SMILES cannot fully utilize the inherent chemical structures and may generate invalid SMILES strings. Therefore, \citep{zheng2019predicting} propose a self-correct transformer to fix the syntax errors of candidate reactants. Recently, G2G \citep{shi2020graph}, RetroXpert \citep{NEURIPS2020_819f46e5} and GraphRetro \citep{somnath2021learning} achieve state-of-the-art performance by decomposing the retrosynthesis into two sub-problems: i) center identification and ii) synthon completion, as shown in Fig.~\ref{fig:global_structure}.  Center identification increases the model scalability by breaking down the target molecule into virtual synthons without utilizing templates. Synthon completion simplifies the complexity of reactant generation by taking synthons as potential starting molecules. For example, RetroXpert and G2G treat it as a SMILES or graph sequence translation problem from synthon to reactant. GraphRetro completes synthons by predicting pre-defined leaving groups, but it does not provide scalable algorithm for attaching leaving groups and can not handle the case of molecule property changes, e.g., there is no residual from $\text{N}$ to $\text{N}^{-}$. Generally, these TF methods are more scalable but perform worse than TB approaches in top-1 accuracy.

\textbf{Challenges} Although the two-step TF framework significantly improves the algorithm's scalability, the overall accuracy is relatively low. A possible solution to this issue is to enhance submodules, i.e., center identification and synthon completion. 1) The current GNN models perform well in top-$1$ accuracy for center identification, but top-$k$ accuracy remains unsatisfactory. \textit{How to develop a more suitable model that provides high top-$k$ accuray} is the first challenge. 2) In addition, synthon completion is the major bottleneck affecting the overall accuracy. Specifically, predicting and attaching residuals for each synthon are difficult because the residual structures could be complex, attaching residuals into synthons may violate chemical rules, and various residuals may agree with the same synthon (e.g., F, CI, Br, and I have similar chemical properties). For researchers, scalability, interpretability, and training efficiency are also important. \textit{How to develop a more accurate, interpretable, and efficient synthon completion model while maintaining the scalability} is the second challenge.

\section{Definition and Overview}
\textbf{Molecule representation} There are two types of dominant representations, i.e., SMILES string \citep{weininger1988smiles} and molecular graph. SMILES is commonly used in early works \citep{liu2017retrosynthetic,schwaller2018found,zheng2019predicting,schwaller2019molecular,tetko2020state} due to its simpleness. Many NLP models can be directly applied to solve related problems in an end-to-end fashion. However, these models cannot guarantee the chemical correctness of the output molecules because they ignore structure information to some extent. Similar to recent breakthroughs \citep{dai2019retrosynthesis,shi2020graph,NEURIPS2020_819f46e5,somnath2021learning}, we take the molecule as a labeled graph $\mathcal{G}(A,X,E)$, where $A$, $X$ and $E$ are adjacency matrix, atom features and bond features, seeing Table.~\ref{tab:symbols}. Under the graph framework, we can effectively apply chemical constraints to ensure the validity of output molecules, which is more controllable and interpretable than SMILES-based approaches.

\begin{table}[H]
    \small
    \centering
    \caption{Commonly used symbols}
    \resizebox{0.9 \columnwidth}{!}{
    \begin{tabular}{ C{0.15 \columnwidth}  C{0.85 \columnwidth}}
    \toprule
         Symbol & Description \\ \midrule
         $\mathcal{G}(A,X,E)$ & Molecular graph with adjacency matrix $A \in \{0,1\}^{n,n}$, atom features $X \in \mathbb{R}^{n,d}$ and bond features $E \in \mathbb{R}^{m,b}$.\\
         $\boldsymbol{x}_{i}$ & The feature vector of atom $i$. $\dim{\boldsymbol{x}_{i}}=d$. \\
         $\boldsymbol{e}_{i,j}$ & The feature vector of bond $(i,j)$. $\dim{\boldsymbol{e}_{i,j}}=b$.\\
         $\mathcal{R}_i,\mathcal{S}_j,\mathcal{P}$ & The $i$-th reactant, the $j$-th synthon and the product.\\
         $c_i$ & $c_i \in \{0,1\}$, indicating whether atom $i$ is the reaction center or not.\\
         $c_{i,j}$ & $c_{i,j} \in \{0,1\}$, indicating whether bond $(i,j)$ is the reaction center or not.\\
    \bottomrule
    \end{tabular}
    }
    \label{tab:symbols}
\end{table}  

\textbf{Problem definition} Retrosynthesis aims to infer the set of reactants $\{\mathcal{R}_i\}_{i=1}^{N}$ that can generate the product $\mathcal{P}$. Formally, that is to learn a mapping function $\mathcal{F}_{\theta}$:

\vspace{-4mm}
\begin{align}
    \label{eq:retrosynthesis}
    \mathcal{F}_{\theta}: \mathcal{P} \mapsto \{\mathcal{R}_i\}_{i=1}^{N_1}.
\end{align}
 
Considering the unknown by-products, the law of conservation of atoms no longer holds here, which makes the problem quite challenging because the algorithm needs to generate new atoms and bonds to get potential reactants.

\textbf{Overview} As shown in Fig.~\ref{fig:global_structure}, we adopt the two-step TF framework due to its scalability and effectiveness. Our method is distinguished from previous works in two folds: 1) We propose a relational graph attention (DRGAT) layer to improve the center identification performance; 2) We use semi-templates and a self-correcting module to facilitate synthon completion, which significantly reduces the problem complexity.

\begin{figure*}[tp]
    \centering
    \includegraphics[width=6.5in]{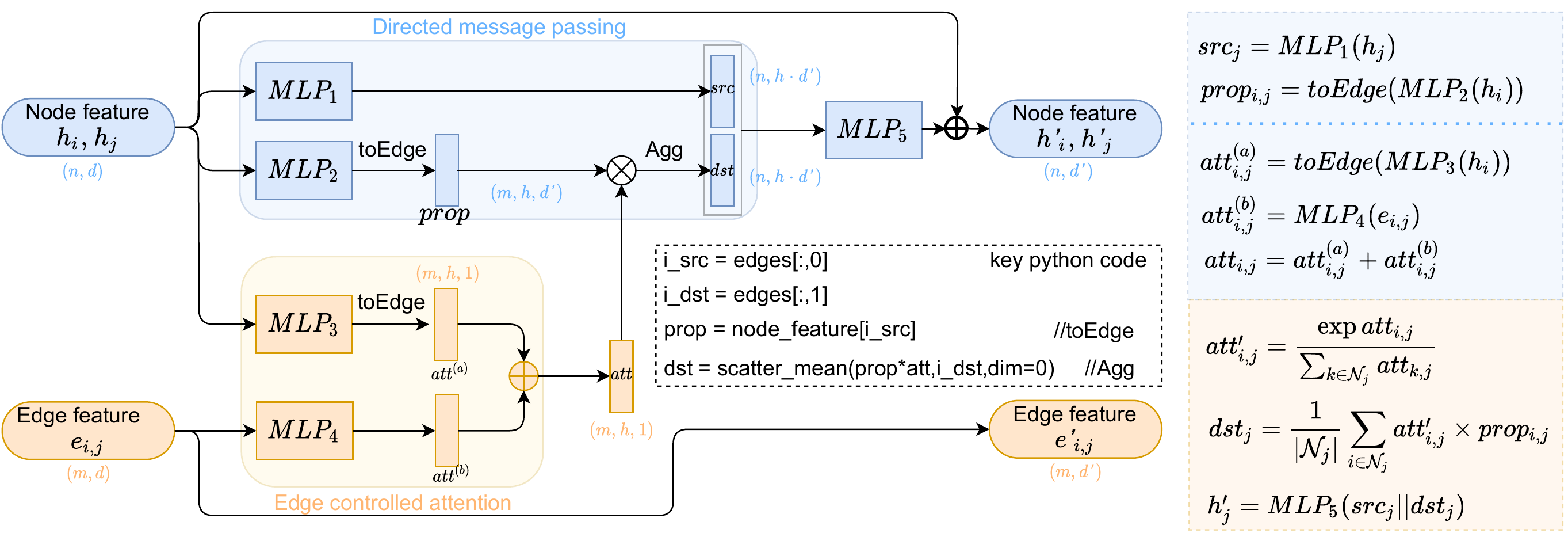}
    \vspace{-4mm}
    \caption{ DRGAT: Directed Relational GAT. DRGAT contains two submodules: directed message passing (DMP) and edge-controlled attention (ECA). DMP uses different MLP to learn features of the source (src) and target (dst) atoms during message passing. ECA utilizes both atom features and bond features to learn the attention weights.}
    \label{fig:DRGAT}
    \vspace{-4mm}
\end{figure*}

\vspace{-2mm}
\section{Methodology}
\vspace{-1mm}
\subsection{Center identification}
Center identification plays a vital role in the two-step retrosynthesis because errors caused by this step directly lead to the final failures. Previous works have limitations, e.g., RetroXpert \citep{NEURIPS2020_819f46e5} provides incomplete prediction without considering atom centers, G2G may leak the edge direction information \citep{shi2020graph}, and GraphRetro \citep{somnath2021learning} provides sub-optimum top-$k$ accuracy. How to obtain comprehensive and accurate center identification results is still worth exploring.

\begin{figure}[h]
    \centering
    \includegraphics[width=3in]{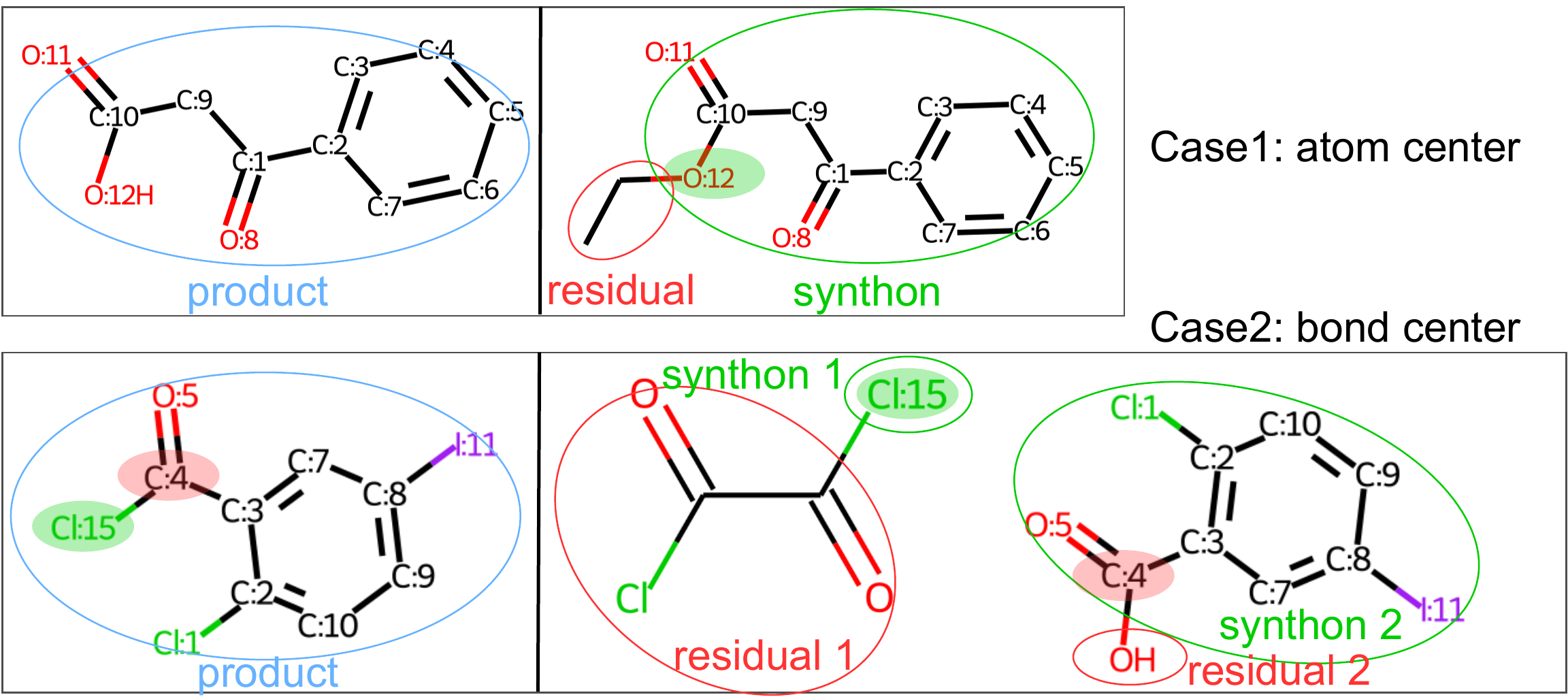}
    \vspace{-3mm}
    \caption{ Reaction centers. Products, reactants, and residuals are circled in blue, green, and red, respectively. We label atoms in reaction centers with solid circles.}
    \label{fig:center_definition}
\end{figure}

\textbf{Reaction centers} We consider both atom centers and bond centers in the product molecule. As shown in Fig.~\ref{fig:center_definition}, from the product to its corresponding reactants, either some atoms add residuals by dehydrogenation without breaking the product structure (case 1), or some bonds are broken to allow new residues to attach (case 2). Both these atoms and bonds are called reaction centers.

\textbf{Directed relational GAT} Commonly used graph neural networks \citep{defferrard2016convolutional,kipf2016semi,velivckovic2017graph} mainly focus on 0 and 1 edges, ignoring edge direction and multiple types, thus failing to capture expressive molecular features. As to molecules, different bonds represent different interatomic interactions, resulting in a multi-relational graph. Meanwhile, atoms at the end of the same bond may gain or lose electrons differently, leading to directionality. Considering these factors, we propose a directed relational graph attention (DRGAT) layer based on the general information propagation framework, as shown in Fig.~\ref{fig:DRGAT}. During message passing, DRGAT extracts source and destination node's features via independent MLPs to consider the bond direction and use the multi-head edge controlled attention mechanism to consider the multi-relational properties. We add shortcut connections from the input to the output in each layer and concatenate hidden representations of all layers to form the final node representation.

\textbf{Labeling and learning reaction centers} We use the same labeling algorithm as G2G to identify ground truth reaction centers, where the core idea is comparing each
pair of atoms in the product $\mathcal{P}$ with that in a reactant $\mathcal{R}_i$. We denote the atom center as $c_i \in \{0,1\}$ and bond center as $c_{i,j} \in \{0,1\}$ in the product $\mathcal{P}$. During the learning process, atoms features $\{ \boldsymbol{h}_i \}_{i=1}^{|\mathcal{P}|}$ are learned from the product $\mathcal{P}$ by applying stacked DRGAT, and the input bond features are $\{\boldsymbol{e}_{i,j}| a_{i,j}=1\}$. Then, we get the representations of atom $i$ and bond $(i,j)$ as

\vspace{-5mm}
\begin{equation}
    \label{eq:center_representation}
    \begin{cases}
        \boldsymbol{\hat{h}}_i = \boldsymbol{h}_i || \mathrm{Mean}(\{\boldsymbol{h}_s\}_{s=1}^{|\mathcal{P}|})  \quad \quad \quad \quad  \textcolor{gray}{\text{// atom}} \\ 
        \boldsymbol{\hat{h}}_{i,j} = \boldsymbol{e}_{ij} || \boldsymbol{h}_i || \boldsymbol{h}_j || \mathrm{Mean}(\{\boldsymbol{h}_s\}_{s=1}^{|\mathcal{P}|}),  \textcolor{gray}{\text{// bond}}
    \end{cases}
\end{equation}
\vspace{-3mm}

% \begin{align}
%     \label{eq:center_representation}
%     \underbrace{ \boldsymbol{\hat{h}}_i = \boldsymbol{h}_i || \mathrm{Readout}(\{\boldsymbol{h}_s\}_{s=1}^{|\mathcal{P}|})}_{\text{atom representation}} \quad \text{and} \\ \underbrace{ \boldsymbol{\hat{h}}_{i,j} = \boldsymbol{e}_{ij} || \boldsymbol{h}_i || \boldsymbol{h}_j || \mathrm{Readout}(\{\boldsymbol{h}_s\}_{s=1}^{|\mathcal{P}|})}_{\text{bond representation}},
% \end{align}

where $\text{Mean}$ and $||$ indicate the average pooling and concatenation operations. Further, we predict the atom center probability $p_i$ and bond center probability $p_{i,j}$ via MLPs:

\vspace{-6mm}
\begin{align}
   \label{eq:center_predict}
   p_i = \mathrm{MLP}_{6}(\boldsymbol{\hat{h}}_i) \quad \text{and} \quad
   p_{i,j} = \mathrm{MLP}_{7}(\boldsymbol{\hat{h}}_{i,j}).
\end{align}
\vspace{-6mm}

Finally, center identification can be reduced to a binary classification, whose loss function is:

\vspace{-3mm}
\begin{equation}
    \begin{aligned}
   \label{eq:center_loss}
   \mathcal{L}_1 = \sum_{\mathcal{P}}( \sum_i { c_i \log{p_i} + (1-c_i) \log{p_i} } + \quad \textcolor{gray}{\text{// atom}} \\
   \sum_{i,j} { c_{i,j} \log{p_{i,j}} + (1-c_{i,j}) \log{p_{i,j}} } ). \quad \textcolor{gray}{\text{// bond}}
    \end{aligned}
\end{equation}

In summary, we propose a directed relational graph attention (DRGAT) layer to learn expressive atom and bond features for accurate center identification prediction. We consider both atom center and bond center to provide comprehensive results. In section.~\ref{sec:result_center_identification}, we show that our method can achieve state-of-the-art accuracy.

\subsection{Synthon completion}
Synthon completion is the main bottleneck of two-step TF retrosynthesis, which is responsible for predicting and attaching residuals for each synthon. This task is challenging because the residual structures could be complex to predict, attaching residuals into synthons may violate chemical rules, and various residuals may agree with the same synthon. Because of these complexities, previous synthon completion approaches are usually inaccurate and cumbersome. Introducing the necessary chemical knowledge to improve interpretability and accuracy can be a promising solution.  However, how to provide attractive scalability and training efficiency is a new challenge.

\begin{figure*}[t]
   \centering
   \includegraphics[width=6.5in]{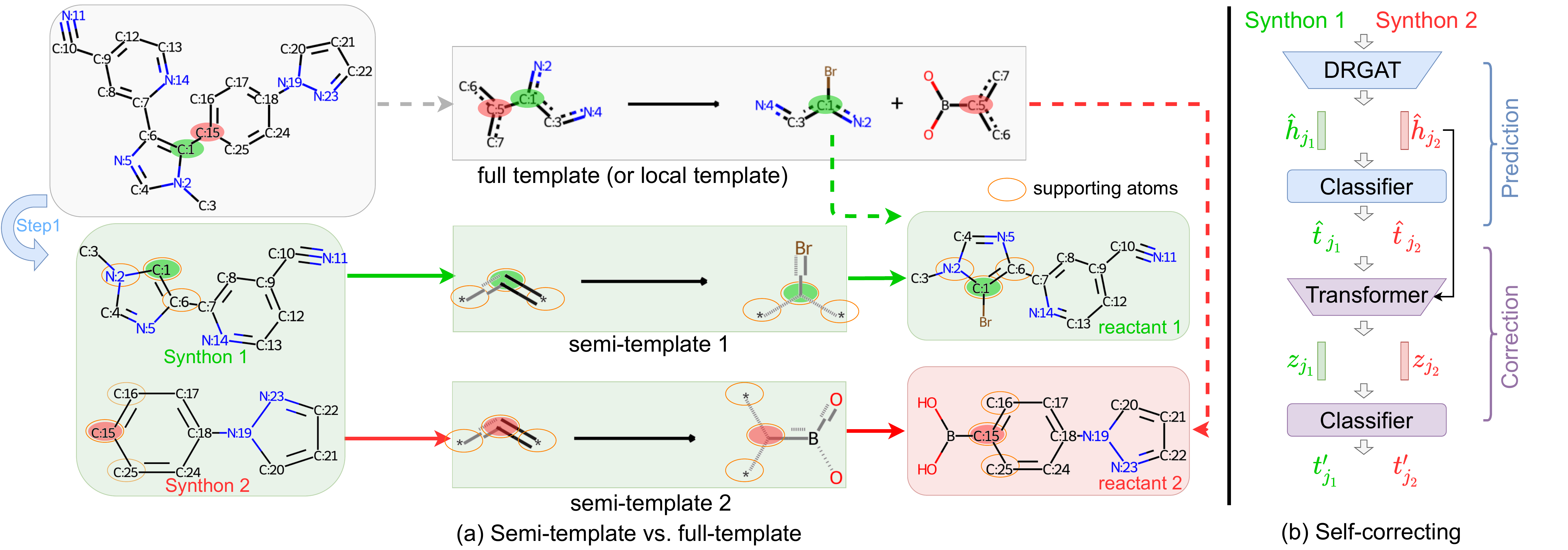}
   \vspace{-3mm}
   \caption{ Predicting semi-template for synthon completion. (a) A full-template can be decomposed into several simpler semi-templates based on synthons. (b) We propose the self-correcting module for more accurate semi-template prediction. }
   \label{fig:semi_tpl}
   \vspace{-3mm}
\end{figure*}

\textbf{Semi-templates} The semi-template used in this paper is the partial reaction pattern of each synthon, seeing Fig.~\ref{fig:semi_tpl}, rather than the whole reaction pattern used in GLN \citep{dai2019retrosynthesis} and LocalRetro \citep{chen2021deep}. Different from GraphRetro \citep{somnath2021learning}, our semi-template encodes the chemical transformation instead of residuals. Similar to the work of forward reaction prediction \citep{segler2016modelling}, semi-template splits a binary reaction into two half reactions. Notably, we use dummy atom $*$ to represent possible synthon atoms that match the semi-template, significantly reducing redundancy. We extract semi-template from each synthon-reactant pair by removing reactant atoms that have exact matches in the synthon. There are two interesting observations: 1) Top-150 semi-templates cover 98.9\% samples; 2) Reactants can be deterministically generated from semi-templates and synthons (introduced later). Based on these observations, synthon completion can be further simplified as a classification problem. In other words, we need to predict the semi-template type for each synthon, and the total number of classes is 150+1. The first 150 classes are top-150 semi-templates, and the 151st class indicates uncovered classes. 

\textbf{Learning semi-templates} For each synthon $\mathcal{S}_j$, denote its semi-template label as $t_j, 1\leq t_j \leq 151$, and the predicted reaction atom set as $\mathcal{C}$. Assume that $\mathcal{\bar{S}}_j$ is the dual synthon of $\mathcal{S}_j$, i.e., $\mathcal{\bar{S}}_j$ and $\mathcal{S}_j$ come from the same product $\mathcal{P}$. We use stacked DRGATs to extract atom features $\{ \boldsymbol{h}_i \}_{i=1}^{|\mathcal{S}_j|}$, $\{ \boldsymbol{\bar{h}}_i \}_{i=1}^{|\mathcal{\bar{S}}_j|}$ and $\{ \boldsymbol{\tilde{h}}_i \}_{i=1}^{|\mathcal{P}|}$. The semi-template representation of $\mathcal{S}_j$ is:

\vspace{-5mm}
\begin{align}
   \label{eq:semi_tpl_representation}
   \boldsymbol{\hat{h}}_j = \text{Mean}(\{ \boldsymbol{h}_i \}_{i \in \mathcal{C}}) || \text{Mean}(\{ \boldsymbol{h}_i \}_{i=1}^{|\mathcal{S}_j|}) || \\ \text{Mean}(\{ \boldsymbol{\bar{h}}_i \}_{i=1}^{|\mathcal{\bar{S}}_j|}) || \text{Mean}(\{ \boldsymbol{\tilde{h}}_i \}_{i=1}^{|\mathcal{P}|}).
\end{align}
\vspace{-5mm}

Based on $\boldsymbol{\hat{h}}_j$, we predict semi-template $\hat{t}_j$ as:

\vspace{-3mm}
\begin{align}
   \label{eq:semi_tpl_pred}
   \hat{t} _j= \argmax_{1\leq c \leq 151}{\tilde{p}_{j,c}}; \quad
   \boldsymbol{\tilde{p}}_j = \text{Softmax}( \mathrm{MLP}_8 (\boldsymbol{\hat{h}}_j) ).
\end{align}
\vspace{-3mm}

Denote $\mathbbm{1}_{\{c\}}(\cdot)$ as the indicator function, the cross-entropy loss used for training is:

\vspace{-3mm}
\begin{align}
   \label{eq:semi_tpl_loss}
   \mathcal{L}_2 = - \sum_{j \in \{1,2,\cdots,|\mathcal{S}_j|\}}  \sum_{1\leq c \leq 151} \mathbbm{1}_{\{c\}}(t_j) \log(\tilde{p}_{j,c}).
\end{align}
\vspace{-3mm}

\textbf{Correcting semi-templates} Considering the pairwise nature of synthons, i.e., dual synthons may contain complementary information that can correct each other's prediction, we propose a self-correcting module to refine the joint prediction results. For $\mathcal{S}_j$, we construct its features as:

\vspace{-3mm}
\begin{align}
   \label{eq:calibrating_representation}
   \boldsymbol{z}_{j} = \boldsymbol{\hat{h}}_{j} || \Phi_{\theta}(\hat{t}_{j}) 
   % \quad and \quad \boldsymbol{\bar{z}}_{j} = \boldsymbol{\hat{\bar{h}}}_{j} || \Phi_{\theta}(\hat{\bar{t}}_{j}),
\end{align}
\vspace{-3mm}

where $\Phi_{\theta}(\hat{t}_{j})$ is the learnable embedding of previous predicted class $\hat{t}_{j}$. Then, we use a multi-layer transformer to capture the interactions between $\boldsymbol{z}_{j}$ and $\boldsymbol{\bar{z}}_{j}$, and get the refined prediction $t'_{j}$:

\vspace{-2mm}
\begin{equation}
   \label{eq:calibrating_representation_updated}
   \begin{cases}
   [\boldsymbol{\hat{z}}_{j}, \boldsymbol{\hat{\bar{z}}}_{j}] = \text{Transformer}([\boldsymbol{z}_{j}, \boldsymbol{\bar{z}}_{j}]) \\
   
   \boldsymbol{p}_j = \text{Softmax}( \mathrm{MLP}_9 (\boldsymbol{\hat{z}}_j) )\\

   t'_j= \argmax_{1\leq c \leq 151}{p_{j,c}}
   \end{cases}
\end{equation}

The correcting loss function is:

\vspace{-5mm}
\begin{align}
   \label{eq:correcting_semi_tpl_loss}
   \mathcal{L}_3 = - \sum_{j \in \{1,2,\cdots,|\mathcal{S}_j|\}}  \sum_{1\leq c \leq 151} \mathbbm{1}_{\{c\}}(t_j) \log(p_{j,c}).
\end{align}
\vspace{-4mm}

In addition, we filter the predicted  pairs based on the prior distribution of the training set. If the prior probability of the predicted pair is zero, we discard the prediction.

\textbf{Applying semi-templates} Once reaction centers, synthons, and corresponding semi-templates are known, we can deduce reactants with almost $100\%$ accuracy. This is not a theoretical claim; We provide a practical residual attachment algorithm in the appendix. 

In summary, we suggest using the semi-templates to improve synthon completion performance with the help of an error mechanism. Firstly, reducing this complex task to a classification problem helps promote training efficiency and accuracy. Secondly, the high coverage of semi-templates significantly enhanced the scalability of TB methods. Thirdly, the deterministic residual attachment algorithm improves interpretability. Fourthly, the proposed self-correcting module can futher improve the prediction accuracy. In section.~\ref{sec:synthon_completion}, we will show the effectiveness of the proposed method.

\vspace{-3mm}
\section{Experiments}
As mentioned earlier, the main contributions of this paper are proposing a DRGAT layer for central identification and suggesting to use a self-correcting semi-template prediction method for synthon completion. The effectiveness of the proposed method is evaluated by systematic experiments, which focus on answering these questions:

\vspace{-3mm}
\begin{itemize}[leftmargin=*]
   \item \textbf{Q1:} For center identification, how much performance gain can be obtained from DRGAT? Where the improvement comes from?
   \vspace{-1mm}
   \item \textbf{Q2:} For synthon completion, can semi-templates reduce template redundancy and improve the synthon completion performance? And how much improvement can be got from the self-correcting mechanism?
   \vspace{-1mm}
   \item \textbf{Q3:} For retrosynthesis, how do we integrate center identification and synthon completion models into a unified retrosynthesis framework? Can SemiRetro outperform existing template-based and template-free methods?
\end{itemize}

\vspace{-3mm}
\subsection{Basic setting}
\textbf{Data} We evaluate SemiRetro on the widely used benchmark dataset USPTO-50k \citep{schneider2016s} to show its effectiveness. USPTO-50k contains 50k atom-mapped reactions with 10 reaction types. Following \citep{dai2019retrosynthesis,shi2020graph,NEURIPS2020_819f46e5}, the training/validation/test splits is 8:1:1. To avoid the information leakage issue \citep{NEURIPS2020_819f46e5, somnath2021learning}, we use canonical SMILES as the original input for both training and testing.

% As mentioned in previous works, the USPTO-50k dataset contains a shortcut in that the product atom with atom-mapping "1" is part of the reaction center in ~75\% of the cases. In our graph-based methods, this shortcut will not be introduced by forbidding atom position encoding.

\textbf{Baselines} Template-based GLN \citep{dai2019retrosynthesis}, template-free G2G \citep{shi2020graph} and RetroXpert \citep{NEURIPS2020_819f46e5} are primary baselines, which not only achieve state-of-the-art performance, but also provide open-source PyTorch code that allows us to verify their effectiveness. To show broad superiority, we also comapre SemiRetro with other baselines, incuding  RetroSim \citep{coley2017computer}, NeuralSym \citep{segler2017neural}, SCROP \citep{zheng2019predicting}, LV-Transformer \citep{chen2019learning}, GraphRetro \citep{somnath2021learning}, MEGAN \citep{sacha2021molecule}, MHNreact \citep{seidl2021modern}, and Dual model \citep{sun2020energy}. As the retrosynthesis task is quite complex, subtle implementation differences or mistakes may cause critical performance fluctuations. We prefer comparing SemiRetro with open-source methods whose results are more reliable. 

% Energy-based approaches \citep{sun2020energy} are ignored because they are more like plug-and-play training strategies and result filters, focusing on enhancing existing models. For simplicity, we leave the use of energy function in the future and concentrate on comparing original retrosynthesis models.

\textbf{Metrics} This paper uses consistent metrics derived from previous literature for both submodule and overall performance. 1). \textit{Center identification}: We report the accuracy of breaking input product into synthons. 2). \textit{Synthon completion}: We present the accuracy of predicing semi-templates from ground truth input synthons. When a product has multiple synthons, the final prediction is correct if and only if all synthons' predicted semi-templates are correct. 3). \textit{Retrosynthesis}: The metric is similar to that of synthon completion, except that the input synthons are also predicted by center identification. In other words, the retrosynthesis is correct if and only if both center identification and synthon completion are correct. Since there may be multiple valid routes for synthesizing a product, we report top-$k$ accuracy. 

\textbf{Implementation details} Thanks to the elegant implementation of G2G \citep{shi2020graph}, we can develop our SemiRetro in a unified PyTorch framework \citep{paszke2019pytorch}, namely TorchDrug. We use the open-source cheminformatics software RDkit \citep{Landrum2016RDKit2016_09_4} to preprocess molecules and SMILES strings. The graph feature extractor consists of 6 stacked DRGAT, with the embedding size 256 for each layer. We train the proposed models for 30 and 50 epochs in center identification and synthon completion with batch size 128 on a single NVIDIA V100 GPU, using the Adam optimizer and OneCycleLR scheduler. We run all experiments three times and report the means of their performance in default. The training costs, atom features, and bond features can be found in the appendix.

% To avoid the label leakage \cite{NEURIPS2020_819f46e5, somnath2021learning}, we ignore atom mapping numbers during the training and evaluation phases.

\subsection{Center identification (\textbf{Q1})}
\label{sec:result_center_identification}
\textbf{A. Objective and setting} This experiment studies \textit{how much center identification performance gain can be obtained from the proposed DRGAT}. Compared to previous works, we use DRGAT to extract graph feature. We trained our model up to 30 epochs, which occupied about 4680 MB of GPU memory, where the batch size is 128, and the learning rate is 1e-3. We point out that we use canonical smiles as inputs and consider both atom center and bond center. In contrast, RetroXpert just considers the bond center and G2G may leak the atomic order information of the non-canonical smiles. For fair and meaningful comparison, we calculate the performance improvement relative to GraphRetro.

\begin{table}[h]
    \small
    \centering
    \begin{tabular}{cccccc}
    \toprule
    &$k=$           & top-1    & top-2    & top-3    & top-5    \\
    \midrule
    \multirow{5}{*}{\rotatebox{90}{known}}     &G2G         & \textcolor{gray}{90.2} & \textcolor{gray}{94.5} & \textcolor{gray}{94.9} & \textcolor{gray}{95.0} \\
                                    &RetroXpert  & \textcolor{gray}{86.0} & --   & --   & --   \\
                                    &GraphRetro  & \underline{84.6} & \underline{92.2} & \underline{93.7} & \underline{94.5} \\
                                    &SemiRetro   & \textbf{86.6} & \textbf{96.7} & \textbf{98.7} & \textbf{99.6} \\ \cline{2-6}
                                    &Improvement & +2.0 & +4.5 & +5.0 & +5.1 \\ \hline

    \multirow{5}{*}{\rotatebox{90}{unknown}}   &G2G   & \textcolor{gray}{75.8} & \textcolor{gray}{83.9} & \textcolor{gray}{85.3} & \textcolor{gray}{85.6}\\
                                    &RetroXpert  & \textcolor{gray}{64.9} & --   & --   & --   \\
                                    &GraphRetro  & \textbf{70.8} & \underline{85.1} & \underline{89.5} & \underline{92.7} \\
                                    &SemiRetro   & \underline{69.3} & \textbf{87.5} & \textbf{93.3} & \textbf{97.9} \\ \cline{2-6}
                                    &Improvement & -- & +2.4 & +3.8 & +5.2 \\ 
    \bottomrule
    \end{tabular}
    \vspace{-1mm}
    \caption{Top-$k$ center identification accuracy when the reaction class is known or unknown. The best and sub-optimum results are highlighted in bold and underline.}
    \label{tab: acc_center}
\end{table}

\textbf{A. Results and analysis} 
1) \textbf{Highest accuracy}: As shown in Table.~\ref{tab: acc_center}, SemiRetro outperforms baselines in most cases with different $k$. 2) \textbf{Better potential}: Since the possible synthesis routes toward a product may be multiple, the top-$k$ accuracy ($k>1$) is important, and the performance gain of SemiRetro rises as $k$ increases, indicating the better potential. In particular, SemiRetro achieves nearly perfect top-5 accuracy on the setting of reaction class known (acc = 99.6\%) and unknown (acc = 97.9\%).
3) \textbf{Attractive efficiency} The proposed model can achieve good performance after training 30 epochs, where each epoch can be finished within 1 minute, see Table.~\ref{tab:train_cost} in the appendix.

% 3) \textbf{Better adaptability}: In a more general and complex case, where the reaction class is unknown, SemiRetro can significantly exceed SOTA methods. For example, SemiRetro outperforms competitors by at least $10\%$ on the top-3 and top-5 accuracy. These results show that the center identification performance has been dramatically improved using DRGAT, which is a good first step towards accurate retrosynthesis prediction.

\textbf{B. Objective and setting} This experiment studies \textit{where the improvement comes from}. Firstly, we compare the DRGAT with well-tuned, off-the-shelf GNNs to show its superiority. Second, we do ablation studies to reveal the really important modules. All models are trained to the 100 epoches to ensure that the different models have converged, both of which have 6 layers GNN and embedding size 256. We present the results here when class is known, and more results can be found in the appendix.

\begin{table}[h]
    \small
    \centering
    \begin{tabular}{ccccc}
    \toprule
    $k=$                     & top-1 & top-2 & top-3 & top-5 \\ \hline
    GCN                    & 82.8      &  94.6     &  97.5     &  99.3     \\
    GAT                    & 75.7      &  88.7     &  92.9     &  96.6     \\
    ChebNet                & 75.0      &  90.5     &  95.3     &  98.3     \\
    GIN                    & 77.0      &  90.8     &  95.2     &  98.3     \\
    RGCN                   & \underline{85.4}      &  \underline{95.9}     &  \underline{98.2}     &  \underline{99.5}     \\ \hline

    DRGAT                  & \textbf{86.6}      &  \textbf{96.7}     &  \textbf{98.7}     &  \textbf{99.6}     \\
    w/o attention          & 86.2      &  96.0     &  98.3     &  99.4     \\
    w/o directed embedding & 86.1      &  96.3     &  98.4     &  99.5     \\
    w/o edge features      & 85.2      &  95.6     &  98.1     &  99.4     \\ 
    \bottomrule
    \end{tabular}
    \caption{Ablation study of center identification, class known.}
    \label{tab: ablation_center}
\end{table}

% \begin{table}[h]
%     \centering
%     \begin{tabular}{ccccc}
%     \toprule
%     $k=$                     & top-1 & top-2 & top-3 & top-5 \\ \hline
%     GCN                    & 65.3      &  84.2     &  91.4     &  96.7     \\
%     GAT                    & 67.3      &  84.6     &  89.9     &  94.7     \\
%     ChebNet                & 67.5      &  84.8     &  95.3     &  98.3     \\
%     GIN                    & 65.0      &  80.3     &  91.0     &  95.6    \\
%     RGCN                   & 69.2      &  86.7     &  93.3     &  97.3     \\ \hline

%     DRGAT                  & 69.3      &  87.5     &  93.3     &  97.9     \\
%     w/o attention          & 68.8      &  86.3     &  92.7     &  97.0     \\
%     w/o directed embedding & 67.1      &  86.6     &  92.9     &  96.8     \\
%     w/o edge features      & 68.6      &  86.6     &  92.3     &  96.7     \\ 
%     \bottomrule
%     \end{tabular}
%     \caption{Ablation study of center identification, class unknown.}
% \end{table}

\textbf{B. Results and analysis}
1) \textbf{Superiority}: From Table.~\ref{tab: ablation_center}, we observe that DRGAT provides better top-$k$ accuracy than existing well-tuned GNNs. 2) \textbf{Ablation}: Both the attention mechanism, the directed embedding, and the usage of edge features bring performance gain.

\begin{table*}[t]
    \small
    \centering
    % \resizebox{2.0 \columnwidth}{!}{
    \begin{tabular}{ C{0.02 \columnwidth} L{0.6 \columnwidth} C{0.065 \columnwidth} C{0.065 \columnwidth} C{0.065 \columnwidth} C{0.065 \columnwidth} C{0.065 \columnwidth} C{0.065 \columnwidth} C{0.065 \columnwidth} C{0.065 \columnwidth} C{0.065 \columnwidth} C{0.065 \columnwidth} } \hline
    \multicolumn{2}{c}{\multirow{3}{*}{\makecell{\\ \\ $k$=}}}          & \multicolumn{8}{c}{top-$k$ accuracy}                                                    \\ \cline{3-10} 
    \multicolumn{2}{c}{}                             & \multicolumn{4}{c}{Reaction class known} & \multicolumn{4}{c}{Reaction class unknown} \\ \cmidrule(lr){3-6} \cmidrule(lr){7-10} 
    \multicolumn{2}{c}{}                             & 1        & 3        & 5        & 10      & 1         & 3        & 5        & 10       \\ \hline
    \multirow{4}{*}{TB} & RetroSim \citep{coley2017computer}       & 52.9     & 73.8     & 81.2     & 88.1    & 37.3      & 54.7     & 63.3     & 74.1     \\
                        & NeuralSym \citep{segler2017neural}     & 55.3     & 76.0     & 81.4     & 85.1    & 44.4      & 65.3     & 72.4     & 78.9     \\
                        & GLN \citep{dai2019retrosynthesis}           & {64.2}     & 79.1     & 85.2     & {90.0}    & {52.5}      & {69.0}     & {75.6}     & {83.7}     \\ \hline
    \multirow{8}{*}{TF} & SCROP \citep{zheng2019predicting}         & 59.0     & 74.8     & 78.1     & 81.1    & 43.7      & 60.0     & 65.2     & 68.7     \\
                        & LV-Transformer \citep{chen2019learning} & --       & --       & --       & --      & 40.5      & 65.1     & 72.8     & 79.4     \\
                        & G2G \citep{shi2020graph}           & 61.0     & 81.3     & 86.0     & 88.7    & 48.9      & 67.6     & 72.5     & 75.5     \\
                        & RetroXpert \citep{NEURIPS2020_819f46e5}    & 62.1     & 75.8     & 78.5     & 80.9    & 50.4      & 61.1     & 62.3     & 63.4     \\
                        & GraphRetro \citep{somnath2021learning}    &  {63.9}     & {81.5}     & {85.2}     & {88.1}    & \underline{53.7}      & {68.3}     & {72.2}     & {75.5}     \\ 
                        & MEGAN \citep{sacha2021molecule}   &  60.7  & \underline{82.0}  & \underline{87.5}  & \underline{91.6}  & 48.1  & 70.7  & 78.4  & \underline{86.1}\\
                        & MHNreact \citep{seidl2021modern} & -- & -- & -- & -- & 50.5 & \underline{73.9} & \textbf{81.0} & \textbf{87.9}\\
                        & Dual \citep{sun2020energy} & \underline{65.7} & 81.9  & 84.7  & 85.9  & 53.6 & 70.7  & 74.6 & 77.0 \\ 
                        \hline
    \multirow{3}{*}{Our}                 & SemiRetro      & \textbf{65.8}    & \textbf{85.7}    & \textbf{89.8}   & \textbf{92.8}   & \textbf{54.9} & \textbf{75.3}   &\underline{80.4}  &{84.1} \\
    & Improvement to GLN & +1.6  & +6.6 & +4.6 & +2.8 & +2.4 & +6.3 & +4.8 & +0.4\\
    & Improvement to G2G & +4.8  & +4.4 & +3.8 & +4.1 & +6.0 & +7.7 & +7.9 & +8.6\\
    \hline

    \end{tabular}
    \caption{Overall performance. The best and sub-optimum results are highlighted in bold and underline. We show the performance gains relative to the important baselines, i.e., template-based GLN and template-free G2G.}
    \vspace{-3mm}
    \label{fig:retrosynthesis}
\end{table*}

\subsection{Synthon completion (\textbf{Q2})}
\label{sec:synthon_completion}
\textbf{A. Objective and setting} This section reveals \textit{the effectiveness of using semi-template} in three folds: 1) reducing the template redundancy, 2) providing good accuracy, and 3) promoting scalability and training efficiency. Firstly, we count the full-templates of GLN and semi-templates introduced in this paper. We visualize the distribution and coverage of top-$k$ templates for analyzing the redundancy. Secondly, we present the accuracy of synthon completion with ground truth synthon inputs. The final reactants are obtained by predicting the semi-templates and applying the residual attachment algorithm. Thirdly, we compare the scalability and training efficiency of different methods in short. We trained our model up to 50 epochs, which occupied about 4108 MB of GPU memory, where the batch size is 128 and the learning rate is 1e-4.

\vspace{-2mm}
\begin{figure}[h]
    \centering
    \includegraphics[width=3in]{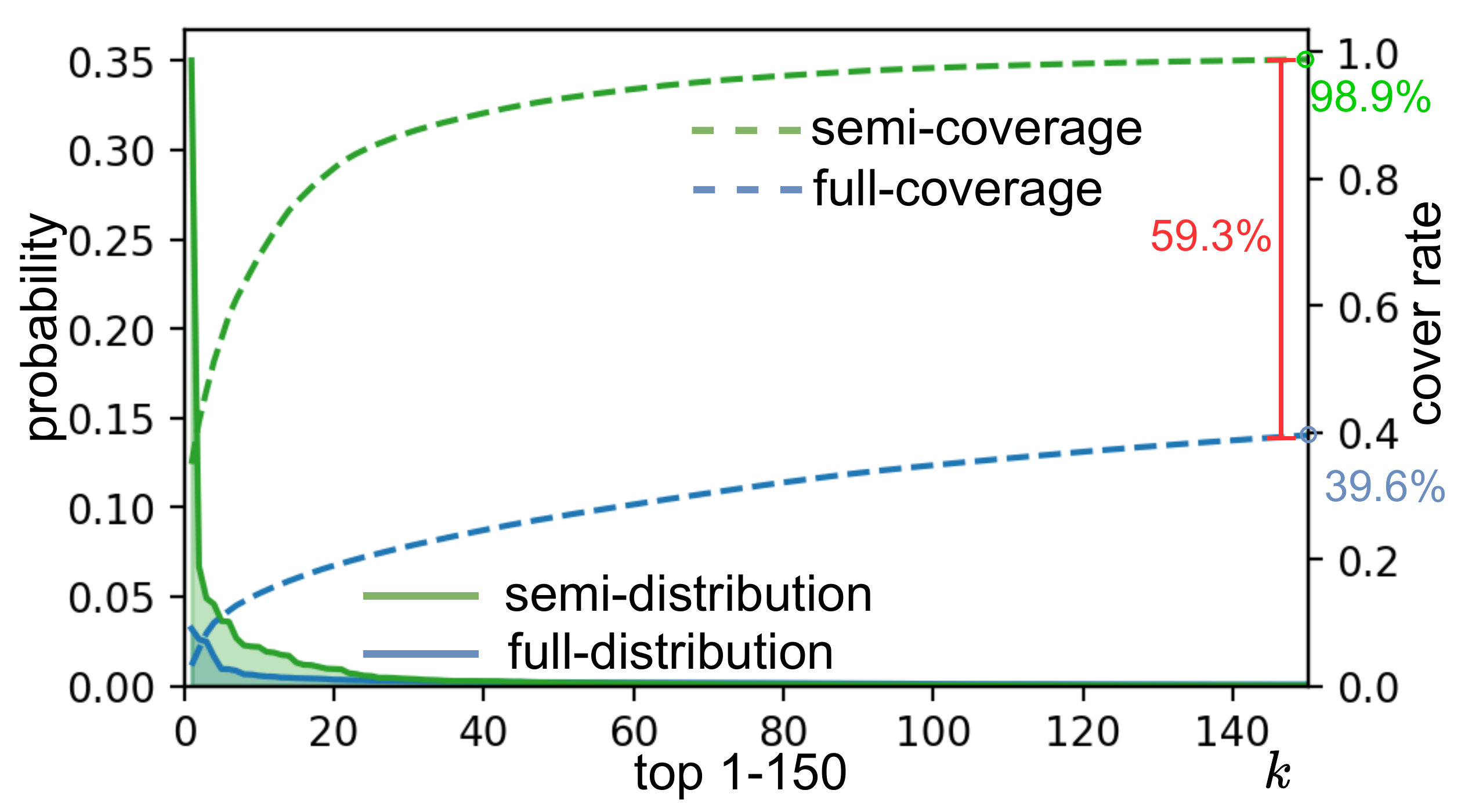}
    \vspace{-4mm}
    \caption{ SemiRetro reduces the template redundancy. }
    \label{fig:redundancy}
\end{figure}
\vspace{-2mm}

% \begin{table}[H]
%     \small
%     \centering
%     \begin{tabular}{ccccc}
%     \toprule
%     $k=$           & top-1    & top-2    & top-3    & top-5    \\
%     \midrule
%     G2G         & 66.8 & -- & 87.2 & 91.5 \\
%     GraphRetro  & 77.4 & 89.6 & 94.2 & 97.6 \\
%     SemiRetro   & 74.7 & 88.9 & 93.6 & 96.3 \\
%     \bottomrule
%     \end{tabular}
%     \vspace{-2mm}
%     \caption{Top-$k$ synthon completion accuracy, class known.}
%     \label{tab: acc_synthon_known}
% \end{table}

% \begin{table}[H]
%     \small
%     \centering
%     \begin{tabular}{ccccc}
%     \toprule
%     $k=$           & top-1    & top-2    & top-3    & top-5    \\
%     \midrule
%     G2G         & 61.1 & -- & 81.5 & 86.7 \\
%     GraphRetro  & 75.6 & 87.7 & 92.9 & 96.3 \\
%     SemiRetro   & 73.1 & 87.6 & 92.6 & 96.0 \\
%     \bottomrule
%     \end{tabular}
%     \vspace{-2mm}
%     \caption{Top-$k$ synthon completion accuracy, class unknown.}
%     \label{tab: acc_synthon_unknown}
% \end{table}

\begin{table}[H]
    \small
    \centering
    \begin{tabular}{cccccc}
    \toprule
    & $k=$           & top-1    & top-2    & top-3    & top-5    \\
    \midrule
    \multirow{3}{*}{\rotatebox{90}{known}}        & G2G         & 66.8 & -- & 87.2 & 91.5 \\
            & GraphRetro  & \textbf{77.4} & \textbf{89.6} & \textbf{94.2} & \textbf{97.6} \\
            & SemiRetro   & \underline{74.7} & \underline{88.9} & \underline{93.6} & \underline{96.3} \\ \cline{2-6}
    \multirow{3}{*}{\rotatebox{90}{unknown}}        & G2G         & 61.1 & -- & 81.5 & 86.7 \\
            & GraphRetro  & \textbf{75.6} & \textbf{87.7} & \textbf{92.9} & \textbf{96.3} \\
            & SemiRetro   & \underline{73.1} & \underline{87.6} & \underline{92.6} & \underline{96.0} \\
    \bottomrule
    \end{tabular}
    \vspace{-2mm}
    \caption{Top-$k$ synthon completion accuracy.}
    \label{tab: acc_synthon}
\end{table}

\textbf{A. Results and analysis} (1) \textbf{Reduce redundancy}: In Fig.~\ref{fig:redundancy}, we show the distribution and coverage of top-$k$ full-templates and semi-templates,
where the former distribution is sharper than the latter, indicating a higher top-$k$ coverage. For example, the top-50 semi-templates cover the case of $92.6\%$, while the full-templates only cover $26.8\%$. Using semi-templates can reduce 11,647 full-templates into 150 semi-templates and increase the cover rate from $93.3\%$ to $98.9\%$. (2) \textbf{ Good accuracy } As shown in Table.~\ref{tab: acc_synthon}, SemiRetro achieves competitive top-$k$ accuracy. (3) \textbf{More scalable and efficient} Although the reported accuracy is not optimum, SemiRetro is more scalable. The semi-template allows encoding property changes of existing atoms and bonds. Moreover, the residual attaching algorithm in the appendix can be used in general cases. In addition, our model can be trained at least 6 times faster than previous synthon completion models such as GLN, G2G, and RetroXpert, seeing the appendix for details.

\textbf{B. Objective and setting} This experiment studies \textit{how much improvement can be got
from the self-correcting mechanism}. We do ablation study by removing the prior distribution based filter and learnable self-correct transformer modules.  We present the results here when class is known, and more results can be found in the appendix.

\begin{table}[h]
    \small
    \centering
    \begin{tabular}{ccccc}
    \hline
    $k=$              & top-1 & top-2 & top-3 & top-5 \\ \hline
    SemiRetro       & 75.0   & 89.4   & 93.9   & 96.7   \\
    w/o filter       & 74.7   & 88.9   & 93.6   & 96.3   \\
    w/o self-correcting \& filter & 71.5  & 87.0  & 92.6  & 96.0  \\ \hline
    \end{tabular}
    \vspace{-2mm}
    \caption{Ablation study of synthon completion, class known.}
    \vspace{-2mm}
    \label{tab: ablation_synthon}
\end{table}

\textbf{B. Results and analysis} From Table.~\ref{tab: ablation_synthon}, we observe that the 
filter improves top-$1$ accuracies about 0.3\%, and the self-correcting transformer improves top-1 accuracy by 3.2\%. This phenomenon shows that the self-correcting mechanism is important to improve accuray.

% \begin{table}[h]
%     \centering
%     \begin{tabular}{ccccc}
%     \hline
%     $k=$              & top-1 & top-2 & top-3 & top-5 \\ \hline
%     SemiRetro       & 73.3   & 88.1   & 93.1   & 96.4   \\
%     w/o filter      & 73.1  & 87.6  & 92.6  & 96.0  \\
%     w/o transformer \& filter &  70.0   &  86.6   & 91.6   & 95.3   \\ \hline
%     \end{tabular}
% \end{table}

\vspace{-2mm}
\subsection{Retrosynthesis (\textbf{Q3})}
\begin{figure}[h]
    \centering
    \includegraphics[width=3.3in]{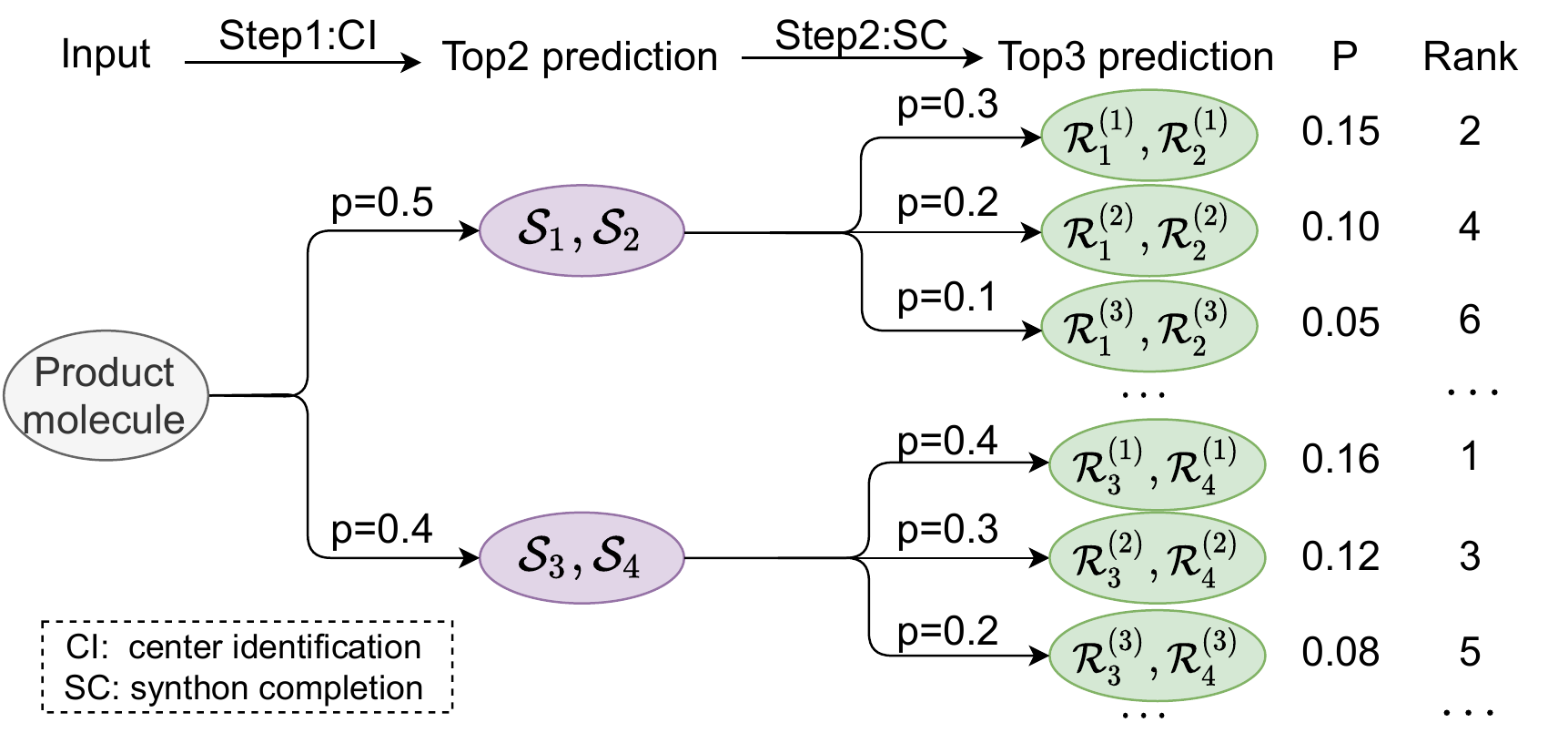}
    \vspace{-4mm}
    \caption{ The retrosynthesis example: combining top-2 CI and top-3 SC to obtain top-6 retrosynthesis results. Note that $\mathcal{S}_i$ indicates the $i$-th synthon, and $\mathcal{R}^{(j)}_i$ is the $j$-th predicted reactant of $\mathcal{S}_i$.}
    \vspace{-4mm}
    \label{fig:inferencez_tree}
\end{figure}

\textbf{A. Objective and setting} We explain \textit{how to combine center identification and synthon completion to provide end-to-end retrosynthesis predictions}. We use a probability tree to search the top-$k$ results, seeing Fig.~\ref{fig:inferencez_tree}, where the probability product of two-step predictions is used to rank these results.

\textbf{B. Objective and setting} We study \textit{whether SemiRetro outperforms existing template-based and template-free methods}. Baseline results are copied from their papers.

\textbf{B. Results and analysis} (1) \textbf{Higher accuracy}: SemiRetro achieves the highest accuracy in most settings, seeing Table.~\ref{fig:retrosynthesis}. As to previous open-source works, template-free G2G and RetroXpert are more scalable than template-based GLN while sacrificing the top-1 accuracy. We use semi-template to reduce the template redundancy and improve the accuracy simultaneously. (2) \textbf{Consistent improvement} While previous methods have their own advantages, they have not yielded such consistent performance gains like SemiRetro under different settings.

\vspace{-2mm}
\section{Conclusion}
We propose SemiRetro for retrosynthesis prediction, which achieves SOTA accuracy and attractive scalability. Specifically, the DRGAT achieves the highest center identification accuracy. The self-correcting semi-template prediction mechanism improves both the accuracy and scalability of synthon completion. Moreover, SemiRetro has favorable training efficiency. We hope this work will promote the development of deep retrosynthesis prediction.

\bibliography{SemiRetro}
\bibliographystyle{icml2021}

% \clearpage
\onecolumn
\appendix
\section{Appendix}

\begin{figure}[h]
   \centering
   \includegraphics[width=5.8in]{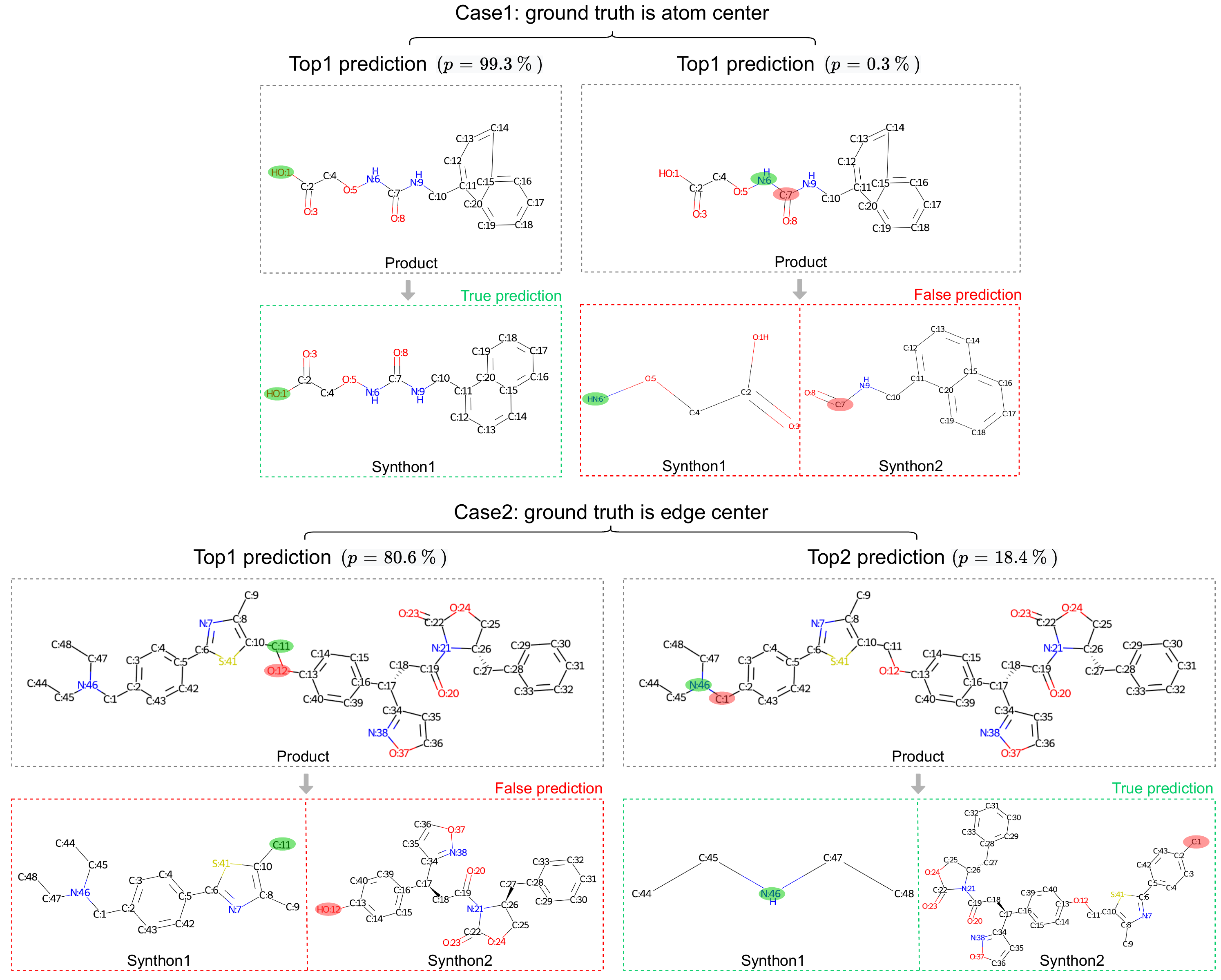}
   \vspace{-4mm}
   \caption{ Visualize results of center identification. Case1: the ground truth is atom center, and the top-1 prediction is correct with the probability $99.3\%$. Case2: The ground truth is edge center, and the top-2 prediction is correct with the probability $18.4\%$.}
   \label{fig:visualize_center_identification}
\end{figure}

\textbf{Center identification} We show top-2 center identification predictions in Fig.~\ref{fig:visualize_center_identification}, where synthons are obtained from breaking edge centers for downstream synthon completion. We present the probability of each prediction where the total probability of top-2 predictions exceeds $98\%$, indicating strong inductive confidence. Since the top-3 predictions are accurate enough, seeing Table.~\ref{tab: acc_center}, we use them for synthon completion.

\vspace{-2mm}

\begin{figure}[H]
   \centering
   \includegraphics[width=5.5in]{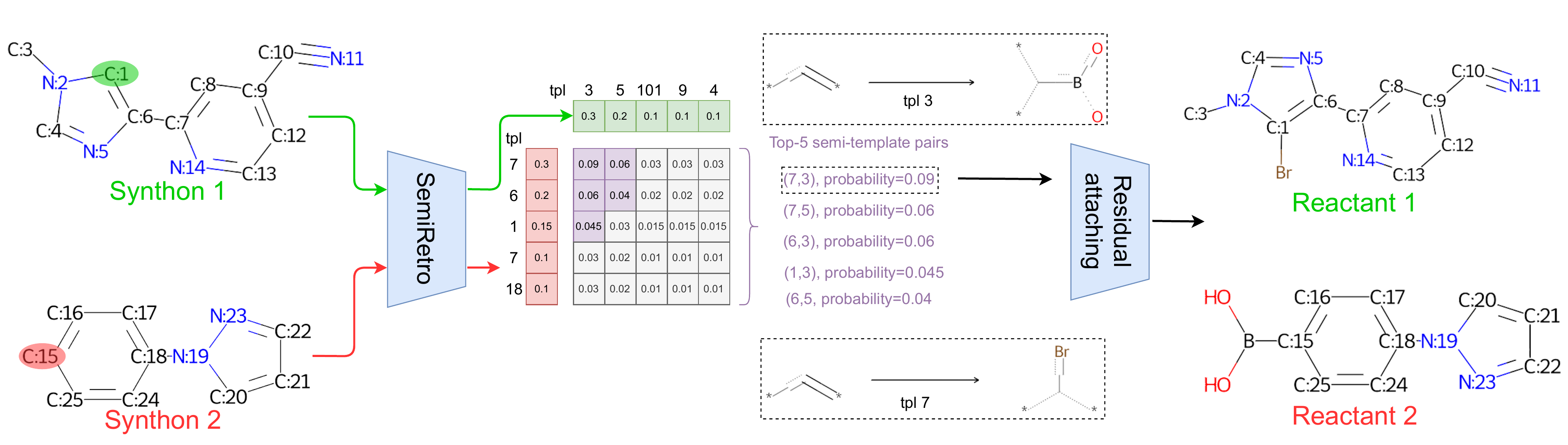}
   \vspace{-4mm}
   \caption{The overall pipeline of synthon completion. The input synthons are the outputs of the center identification module, coming from the same product molecule. We get the top-5 semi-template predictions and their probabilities of each synthon using SemiRetro (synthon completion network), then generate the joint distribution of semi-templates. We choose the top-5 predictions from this joint distribution and apply the residual attachment algorithm (introduced later) to get the final reactants.}
   \label{fig:visual_synthon_completion}
\end{figure}

\vspace{-2mm}
\textbf{Synthon completion} In Fig.~\ref{fig:visual_synthon_completion}, we present the process of predicting multiple reactants of the same product. This process provides an end-to-end view of synthon completion, containing semi-template prediction, top-$k$ results search, and semi-template application. By default, we choose the top-4 synthon completion results for each center identification output as part of the final top-10 retrosynthesis results. 

\vspace{-2mm}

\clearpage

\begin{table}[h]
   \centering
   \caption{ Residual attachment algorithm. For easy and quick understanding, we demonstrate the core idea by visual samples. The detailed implementation can be found in the open-source code. }
   \label{tab:residual_attach}

   \includegraphics[width=5.5in]{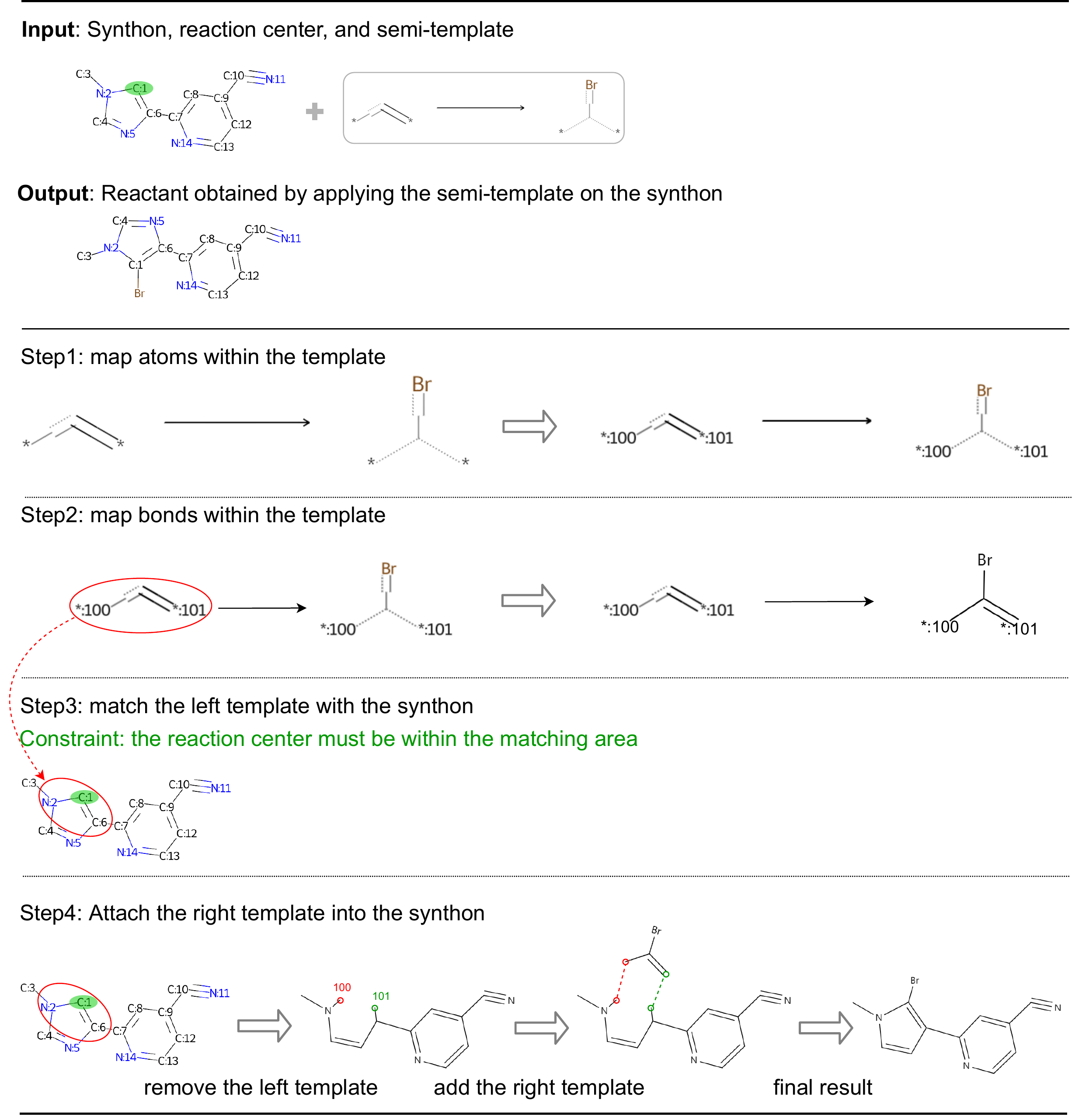}
\end{table}

\textbf{Platform} The platform for our experiment is ubuntu 18.04, with a Intel® Xeon® Gold 6240R Processor and 256GB memory. We use a single NVIDIA V100 to train models, where the CUDA version is 10.2.

\begin{table}[H]
   \centering
   \begin{tabular}{cccccccc}
   \toprule
   & Retrosynthesis   & \multicolumn{3}{c}{Center identification} & \multicolumn{3}{c}{Synthon completion} \\ \cmidrule(lr){2-2} \cmidrule(lr){3-5} \cmidrule(lr){6-8}
   & GLN & RetroXpert      & G2G     & SemiRetro     & RetroXpert     & G2G    & SemiRetro    \\
   \midrule
   time/epoch    & 785s    &    440s        &  58s       &    \textbf{56s}           &  330s              & 322s  & \textbf{33s} \\ % 4108MB synthon completion
   GPU memory/sample & 274.7MB    & 55.7MB   & 46.1MB    & \textbf{36.6MB}     & 147.7MB      & 65.7MB & \textbf{38.2MB}  \\
   total epochs & 50  & 80 & 100 & 30  & 300 & 100 & 50\\
   \bottomrule
   \end{tabular}
   \caption{The training costs of different methods. We run the open-source code of these methods on the same platform, reporting the training time per epoch and occupied GPU memory per sample.  We also show the total training epochs mentioned in their paper (preferred) or code. If the author reports training steps, we calculate $\mathrm{epochs_{total} = steps_{total}/steps_{interval}}$. }
   \label{tab:train_cost}
\end{table}

\clearpage
\paragraph{Important details} We follow the setting of G2Gs, which provides open-source code on https://github.com/DeepGraphLearning/torchdrug/. G2Gs use different atom features in their open-source code for center identification and synthon completion. We have also tried to combine all these atom features and use the same set of features in center identification and synthon completion models. The combined atom features do not make a significant difference. In this paper, we use the same feature for both center identification and synthon completion.

\begin{table}[h]
   \centering
   \begin{tabular}{cc}
   \toprule
   Name        & Description \\
   \midrule
   Atom type   & Type of atom (ex. C, N, O), by atomic number                                     \\
   \# Hs       & one-hot embedding for the total number of Hs (explicit and implicit) on the atom \\
   Degree      & one-hot embedding for the degree of the atom in the molecule including Hs        \\
   Valence     & one-hot embedding for the total valence (explicit + implicit) of the atom        \\
   Aromaticity & Whether this atom is part of an aromatic system.                                 \\
   Ring      & whether the atom is in a ring                                                    \\
   Ring 3    & whether the atom is in a ring of size 3                                          \\
   Ring 4    & whether the atom is in a ring of size 4                                          \\
   Ring 5    & whether the atom is in a ring of size 5                                          \\
   Ring 6    & whether the atom is in a ring of size 6                                          \\
   Ring 6+   & whether the atom is in a ring of size larger than 6\\
   \bottomrule
   \end{tabular}
   \caption{Atom features for center identification and synthon completion.}
   \label{tab:atom_feature_center}
\end{table}

\begin{table}[h]
   \centering
   \begin{tabular}{cc}
   \toprule
   Name           & Description\\
   \midrule
   Bond type      & one-hot embedding for the type of the bond                 \\
   Bond direction & one-hot embedding for the direction of the bond            \\
   Stereo         & one-hot embedding for the stereo configuration of the bond \\
   Conjugation    & whether the bond is considered to be conjugated            \\
   Bond length    & the length of the bond\\
   \bottomrule
   \end{tabular}
   \caption{Bond features for center identification.}
   \label{tab:bond_feature}
\end{table}

\end{document}